\documentclass[10pt,twocolumn,letterpaper]{article}

\usepackage{iccv}
\usepackage{times}
\usepackage{epsfig}
\usepackage{graphicx}
\usepackage{amsmath}
\usepackage{amssymb}

\usepackage{booktabs}
\usepackage{multirow}
\usepackage{multicol}
\usepackage{times}
\usepackage{mathrsfs}
\usepackage[font=small]{caption}
\usepackage{subcaption}
\usepackage{enumitem}
\usepackage{color}
\usepackage[table]{xcolor}

\usepackage{pifont}
\newcommand{\hollowstar}{\text{\ding{73}}}

\DeclareMathOperator*{\argmax}{arg\,max}

\usepackage[pagebackref=true,breaklinks=true,colorlinks,bookmarks=false]{hyperref}

\iccvfinalcopy 

\def\iccvPaperID{7710} 

\ificcvfinal\fi

\begin{document}

\title{Elaborative Rehearsal for Zero-shot Action Recognition}

\author{Shizhe Chen\thanks{This work was performed when Shizhe Chen was at Carnegie Mellon University.}\\
Inria, France\\
{\tt\small shizhe.chen@inria.fr}
\and
Dong Huang\\
Carnegie Mellon University, USA\\
{\tt\small donghuang@cmu.edu}
}

\maketitle
\ificcvfinal\thispagestyle{empty}\fi

\begin{abstract}
The growing number of action classes has posed a new challenge for video understanding, making Zero-Shot Action Recognition (ZSAR) a thriving direction. The ZSAR task aims to recognize target (unseen) actions without training examples by leveraging semantic representations to bridge seen and unseen actions. However, due to the complexity and diversity of actions, it remains challenging to semantically represent action classes and transfer knowledge from seen data. In this work, we propose an ER-enhanced ZSAR model inspired by an effective human memory technique Elaborative Rehearsal (ER), which involves elaborating a new concept and relating it to known concepts. Specifically, we expand each action class as an \textbf{Elaborative Description} (ED) sentence, which is more discriminative than a class name and less costly than manual-defined attributes. Besides directly aligning class semantics with videos, we incorporate objects from the video as \textbf{Elaborative Concepts} (EC) to improve video semantics and generalization from seen actions to unseen actions. Our ER-enhanced ZSAR model achieves state-of-the-art results on three existing benchmarks. Moreover, we propose a new ZSAR evaluation protocol on the Kinetics dataset to overcome limitations of current benchmarks and first compare with few-shot learning baselines on this more realistic setting. 
Our codes and collected EDs are released at \url{https://github.com/DeLightCMU/ElaborativeRehearsal}.
\end{abstract}

\section{Introduction}

Supervised video action recognition (AR) has made great progress in recent years, benefited from new models such as 3D convolutional neural networks~\cite{feichtenhofer2020x3d,feichtenhofer2019slowfast,tran2018closer} and large-scale video datasets \cite{carreira2017quo,ghadiyaram2019large}.
These supervised models require abundant training data for each action class.
However, desired action classes are continuously increasing with the explosive growth of video applications on smart phones, surveillance cameras and drones.
It is prohibitively expensive to collect annotated videos for each action class to fuel the training needs of existing supervised models.
In order to alleviate such burden, Zero-Short Action Recognition (ZSAR) \cite{xu2017transductive} has become a thriving research direction, which aims at generalizing AR models to unseen actions without using any labeled training data of unseen classes.

\begin{figure} 
	\centering
	\includegraphics[width=1\linewidth]{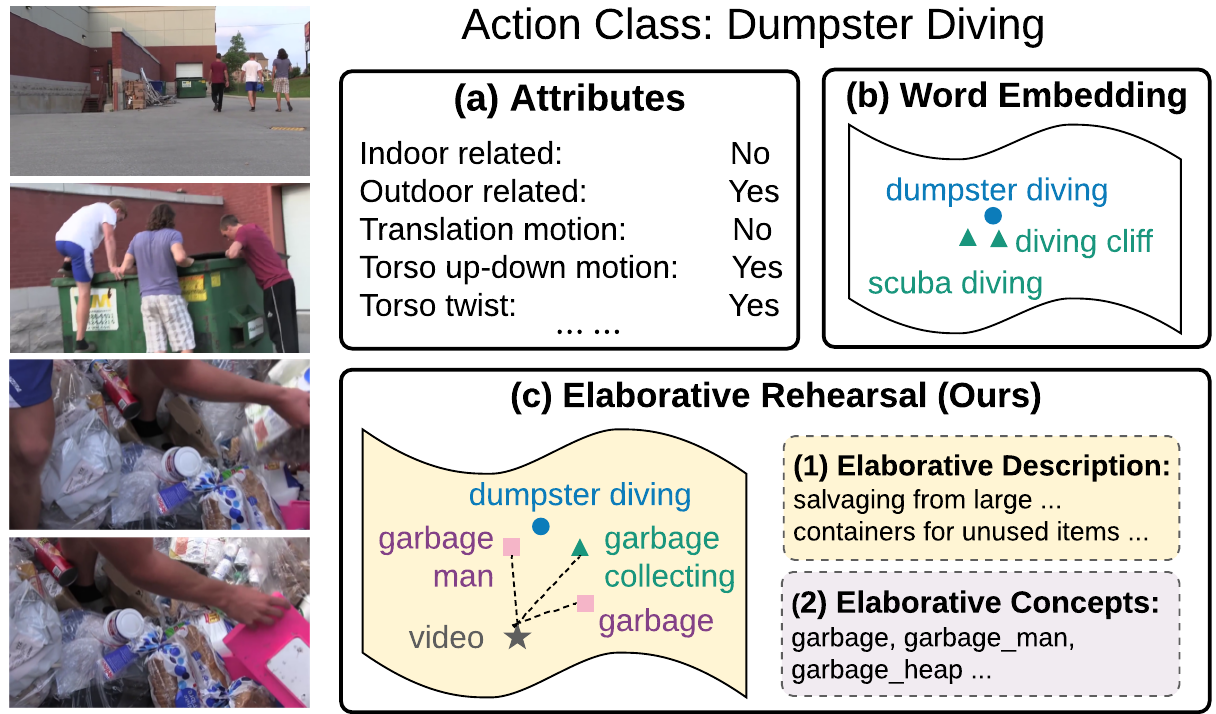}
	\caption{Attributes and word embeddings are insufficient to semantically represent action classes. Our Elaborative Rehearsal approach defines actions by Elaborative Descriptions (EDs) and associates videos with Elaborative Concepts (ECs, known concepts detected from the video), which improve video semantics and generalization video-action association for ZSAR. (\hollowstar for videos, $\triangle$ for seen actions, $\circ$ for unseen actions, and $\square$ for ECs)}
	\label{fig:intro}
\end{figure}

A common approach for ZSAR is to embed videos and action classes into a joint semantic space \cite{frome2013devise,xu2015semantic}, so that the associations between video and seen actions can be transferred to unseen actions.
However, \emph{how to semantically represent action classes for above associations} is a challenging problem due to the complexity and diversity of actions.
As shown in Figure~\ref{fig:intro}(a), early works employ manual-defined attributes \cite{liu2011recognizing} to represent actions. Despite being a natural methodology, it is hard and expensive to define a complete set of atom attributes that generalizes to arbitrary actions. 
To overcome difficulties in attribute definition, recent works adopt word embeddings of action names \cite{xu2017transductive,brattoli2020rethinking} as class semantic representations. Though simple and effective, word embeddings can be ambiguous. Words have different meanings in different context and some actions might not even be interpreted literally according to their names such as the ``dumpster diving'' action in Figure~\ref{fig:intro}(b), which are confusing to relating different action classes.

In addition to class semantic representations of actions, it has been under-explored in existing ZSAR works on \emph{how to learn powerful and generalizable video semantic representations}.
Only until recently, deep features \cite{he2016deep,tran2015learning} have been used to overtake traditional hand-crafted features such as fisher vectors of improved dense trajectory descriptors \cite{wang2013action,xu2017transductive}.
One line of work~\cite{gao2019know,jain2015objects2action} utilizes objects recognized by deep image networks as video descriptors, which assumes that object recognition in image domain is prior knowledge for more advanced action recognition. The predicted objects are naturally embedded in the semantic space and thus can be well generalized to recognize actions even without any video example \cite{jain2015objects2action}. However, the video is more than collections of objects, but contains specific relationships among objects. Therefore, it is insufficient to represent video contents purely using object semantics. Another direction of works \cite{brattoli2020rethinking}, instead, directly employs state-of-the-art 
video classification networks in ZSAR. Though powerful enough to capture spatio-temporal information in the video, they are prone to overfit on seen action classes and transfer poorly to unseen ones.

In this work, we take inspiration from a well-established human memory technique, namely Elaborative Rehearsal (ER) \cite{benjamin2000relationship}, for ZSAR.
When we learn a new item such as ``dumpster diving'', we first expand the phrase into a readily comprehensible definition, and then relate the definition to known information in our long-term memory, thereby fostering retention of the item.
In a similar manner, we propose an ER-enhanced model to generalize AR models for new actions. Our approach advances ZSAR in three main aspects under the common paradigm of joint semantic space learning~\cite{frome2013devise,xu2015semantic}:
\textbf{(1)} For the class semantic representation of actions, we construct Elaborative Descriptions (ED) from class names to comprehensively define action classes as shown in 
Figure~\ref{fig:intro}(c), and embed the ED leveraging prior knowledge from pre-trained language models.
\textbf{(2)} For the video semantic representation, we propose two encoding network streams that jointly embed spatio-temporal dynamics and objects in videos. 
We use a pre-trained image object classification model \cite{kolesnikov2019big} to generate the Elaborative Concepts (EC) of objects. Since it is highly likely that 
some common objects involved in seen and unseen classes, incorporating EC in video semantics improves the generalization on unseen classes. 
\textbf{(3)} To further improve generalization of video semantic representations, we propose an ER objective to enforce the model to rehearse video contents with additional semantic knowledge from EC. 
The embedding of EC shares the same embedding function as the ED of action classes, which also implicitly makes our ZSAR model more generalizable to diverse class semantic representation. 
Our ER-enhanced ZSAR model achieves state-of-the-art performance on the widely used benchmarks including Olympic Sports~\cite{niebles2010modeling}, HMDB51~\cite{kuehne2011hmdb} and UCF101~\cite{soomro2012ucf101} datasets.

Moreover, existing ZSAR benchmarks are relative small and contain overlapped classes with video datasets for feature training. In order to benchmark ZSAR progress on a more realistic scenario, we further propose a new ZSAR evaluation protocol based on a large-scale supervised action dataset Kinetics \cite{carreira2018short,carreira2017quo}.
In our Kinetics ZSAR benchmark, we demonstrate the first case where ZSAR performance is comparable to a simple but strong few-shot learning baseline under clear split of seen and unseen action classes.



\section{Related Work}

\begin{table*}
	\centering
	\small
	\begin{tabular}{l|l|l|l} \toprule
		&  Class Name & Elaborative Description (ED) & ED Source \\  \midrule
		\multirow{4}{*}{Action Class} & cleaning gutters & \begin{tabular}[c]{@{}l@{}}cleaning gutters : make clean ; remove dirt , marks , or stains from . a shallow \\ trough fixed beneath the edge of a roof for carrying off rainwater .\end{tabular} & \multirow{4}{*}{\begin{tabular}[c]{@{}l@{}}Wikipedia +\\ Dictionary +\\ Modification\end{tabular}} \\ \cmidrule{2-3}
		& clean and jerk & \begin{tabular}[c]{@{}l@{}}clean and jerk : a two - movement weightlifting exercise in which a weight is \\ raised above the head following an initial lift to shoulder level .\end{tabular} &  \\ 
		\midrule
		Object Concept & chipboard & \begin{tabular}[c]{@{}l@{}}chipboard : a cheap hard material made from wood chips that are pressed together \\ and bound with synthetic resin\end{tabular} & WordNet \\\bottomrule
	\end{tabular}
	\caption{Examples of Elaborative Descriptions (ED) for action classes and object concepts.}
	\label{tab:label_defn_examples}
\end{table*}


\noindent\textbf{Supervised Action Recognition.}
The rapid development of deep learning \cite{goodfellow2016deep} has vigorously promoted AR research.
Early deep models \cite{karpathy2014large,simonyan2014two,wang2018temporal} adopt 2D convolutional neural networks (CNNs) in temporal domain.
To more effectively encode temporal dynamics in videos, 3D CNNs \cite{tran2015learning} are proposed but are computation and parameter heavy, which require large-scale datasets to train.
Therefore, different approaches have emerged to improve 3D CNNs.
Carreira \etal \cite{carreira2017quo} propose I3D network which inflates 2D CNN to 3D CNN to learn spatio-temporal features.
Tran \etal \cite{tran2018closer} and Qiu \etal \cite{qiu2017learning} decompose 3D convolution into 2D spatial and 1D temporal convolutions.
Wang \etal \cite{wang2018non} insert non-local blocks into 3D CNNs to capture long-range dependencies.
Feichtenhofer \etal \cite{feichtenhofer2019slowfast} introduce slowfast network with two pathways operating at different frame rates, and further explores expansion of 2D CNNs along space, time, width and depth in \cite{feichtenhofer2020x3d}.
Lin \etal \cite{lin2019tsm} propose temporal shift module (TSM) to achieve temporal modeling at 2D computational costs and parameters.
Despite strong performance, these supervised models cannot recognize new classes without training examples. In this work, we generalize the AR models to recognize unseen actions.

\vspace{0.2em}
\noindent\textbf{Zero Shot Learning.}
Most ZSL works \cite{akata2015label,akata2015evaluation,frome2013devise,wang2018zero,xian2018zero,zhang2017learning} focus on the image domain to recognize unseen objects. A comprehensive survey can be found in \cite{xian2018zero}. Here we mainly review joint semantic space based methods.
ALE \cite{akata2015label}, DEVISE \cite{frome2013devise} and SJE \cite{akata2015evaluation} use bilinear compatibility function to associate visual and class representations, with different objectives for training.
ESZSL \cite{romera2015embarrassingly} proposes an objective function with closed form solution for linear projection.
DEM \cite{zhang2017learning} proposes to use visual space as embedding space to address hubness problem in ZSL.
Different from above approaches, Wang \etal \cite{wang2018zero} predict classification weights based on knowledge graphs of classes.
Except using different features, the ZSL methods in image domain can be applied for zero-shot action recognition.

\vspace{0.2em}
\noindent\textbf{Zero Shot Action Recognition.}
As the main focus of our work is to learn better video and action semantic representations for ZSAR, we group existing works according to types of semantic representation of actions.
The first type takes manual-defined attributes \cite{liu2011recognizing,zellers2017zero} to represent an action. Gan \etal \cite{gan2016learning} improve attribute detection via multi-source domain generalization.
Nevertheless, the attributes of actions are harder to define compared with the image counterparts.
The second type then exploits objects as attributes.
Jain \etal \cite{jain2015objects2action} detect objects in videos and associate videos to action category with maximum object similarity.
Gan \etal \cite{gan2016concepts} propose to select discriminative concepts.
Gao \etal \cite{gao2019know} utilize graph networks to learn action-object relationships and then match objects in the video with action prototypes.
Though effective, the above work ignore spatio-temporal relationships in videos and actions.
The third type of approaches uses word embedding of action names \cite{brattoli2020rethinking,mandal2019out,qin2017zero,xu2017transductive} as semantic representation.
Qin \etal \cite{qin2017zero} derive error-correcting output codes for actions via both category-level embedding and intrinsic data structures.
The recent work \cite{brattoli2020rethinking} argues that end-to-end training is important for ZSAR and proposes to train a 3D CNN to predict word embedding of action names.
However, word embeddings can be ambiguous and mislead knowledge transfer among action classes.
The most similar work to ours is \cite{wang2017alternative}, which employs texts and images as alternative semantic representation for actions, but their text descriptions are rather noisy and inferior to attributes or word embeddings.


\section{Our Approach}

\begin{figure*}
	\centering
	\includegraphics[width=1\linewidth]{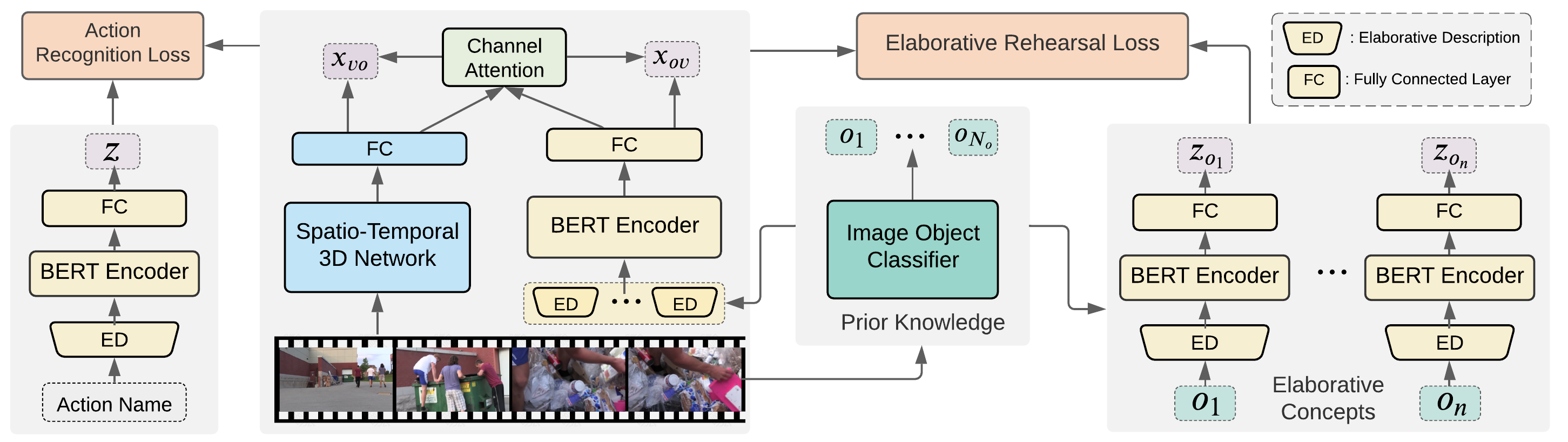}
	\caption{Architecture of our ER-enhanced ZSAR model. The action class embedding function (left) extends action names to EDs towards action class embedding $z$. The multimodal video embedding function generates spatio-temporal and object features $[x_{vo},x_{ov}]$ (middle) for the video. The ER loss utilizes recognized object semantics $z_{o}$ (right) to match $[x_{vo},x_{ov}]$ which improves the action recognition loss.}
	\label{fig:framework}
\end{figure*}

In ZSAR, we are given a source dataset $\mathcal{D}^s=\{(v^n, y^n)\}_{n=1}^N$ of $N$ video with labels from seen action classes $\mathcal{S}=\{1, \cdots, S\}$, 
where $v^n$ is a video clip and $y^n \in \mathcal{S}$ is the label.
$\mathcal{D}^{t}=\{(v^m)\}_{m=1+N}^{N+M}$ is the target dataset of $M$ videos with labels from unseen action classes $\mathcal{T}=\{S+1, \cdots, S+T\}$.
The goal of ZSAR is to classify $v^m \in \mathcal{D}^t$ over unseen classes $\mathcal{T}$ with AR models only trained on $\mathcal{D}^s$.
Following \cite{frome2013devise,xu2015semantic}, the main architecture of our ZSAR model embedes videos and action classes (texts) into a joint semantic space, in which classes of similar semantics are closer as nearest-neighbors. Their respective embedding functions are a video embedding function $\phi(v)$ and an action class embedding function $\psi(y)$. Both functions are only trained on $\mathcal{D}^s$, and to be tested on $\mathcal{D}^{t}$. 

In the rest of the section, we present the novel components of our ER-enhanced ZSAR model: Elaborative Description (ED), action class embedding function $\psi(y)$, video embedding function $\phi(v)$, and Elaborative Rehearsal (ER) loss. The framework is illustrated in Figure~\ref{fig:framework}.


\subsection{Elaborative Description (ED)}
\label{sec:elaborative_description}

For each action class name or each object concept, we concatenate a name and its sentence-based definition as an ED.
Examples of EDs are listed in Table~\ref{tab:label_defn_examples}, which are more discriminative than class names and easier to generate than attributes to semantically represent an action or an object.

\vspace{0.2em}
\noindent\textbf{Justification for Human Involvement.}
ZSL demands class-wise semantic representations, which might involve human to construct, but costs significantly less than sample-wise annotation efforts in supervised training.
In fact, it is a vital step of ZSL to design a high-quality semantic representation with less class-wise annotating efforts.
For ZSL on general object classification task \cite{frome2013devise,li2017zero,xian2018zero}, word embeddings of class names are gaining popularity as semantic representation, because the semantic embeddings of general object words are well-learned in pre-trained language models and can be used as prior knowledge.
However, word embeddings are not applicable to other domains such as fine-grained ZSL for bird species \cite{huynh2020fine} where the class name provides little information about visual appearances. Manual-defined attributes \cite{huynh2020fine} or cleaned text descriptions \cite{paz2020zest} are required in such scenarios.
The situation is similar in ZSAR, where action names alone are not discriminative enough to represent context of the action. For example, the action ``fidgeting'' in the Kinetics dataset~\cite{carreira2017quo} denotes ``playing fidget spinner'' instead of its common meaning of ``making small movements''.
Therefore, it is necessary for human involvement to clarify the action definitions.
Compared to carefully designed and annotated attributes, a more natural way for we humans is to describe the visual process of target actions in natural language, which motivates us to collect sentence-based ED for action class representation.

\vspace{0.2em}
\noindent\textbf{Construction of Elaborative Description.}
Defining actions is more complicated than objects. In the ImageNet dataset \cite{deng2009imagenet}, object classes are directly linked to concepts in WordNet \cite{miller1995wordnet}, and thus EDs of objects are straightforward to obtain. However, currently there are no such resources to define actions.
To reduce manual efforts of writing EDs from scratch, we first automatically crawl candidate sentences from Wikipedia and dictionaries using action names as queries. Then we ask annotators to select and modify a minimum set of sentences from the candidates to describe the target action given few video exemplars.
More details are presented in the supplementary material.
It takes less than 20 seconds on average to generate the ED per action class, which is very efficient.
The average length of EDs for actions in the Kinetics dataset \cite{carreira2017quo} is 36 words.

\subsection{Action Class Embedding $\psi(y)$}
\label{sec:action_class_function}

Denote $d=\{w_1, \cdots, w_{N_d}\}$ the ED for action $y$, where $w_i$ is the composed word. The goal of action class embedding is to encode $d$ into semantic feature $z \in \mathbb{R}^{K}$ with dimension of $K$.

In order to capture the sequential order in $d$ and transfer prior knowledge from large-scale text models, unlike previous works that use tf-idf \cite{qiao2016less}, average word embedding \cite{xu2017transductive} or RNNs trained from scratch \cite{zhang2017learning}, we propose to employ a pre-trained BERT model \cite{devlin2018bert} for ED encoding.
The BERT model has demonstrated excellent capability in implicitly encoding commonsense knowledge \cite{cui2020does}, which is beneficial to embed global semantics of the sentences.

Denote $h_i \in \mathbb{R}^{768}$ as the hidden state from the last layer of BERT for word $w_i$, we apply average pooling to obtain a sentence-level feature $\bar{h}$:
\begin{equation}
	\bar{h} = \frac{1}{N_d}\sum\nolimits_{i=1}^{N_d} h_i.
\end{equation}
Since there are multi-layers of self-attention in BERT, the content words are more strengthened than other stopwords. Therefore, we did not observe performance gains using more complicated methods to aggregate $h_i$ than our average pooling.
Then we use a linear transformation layer to translate $\bar{h}$ into the joint semantic embedding space:
\begin{equation}
\hat{z} = W_c \bar{h} + b_c,
\end{equation}
where $W_c \in \mathbb{R}^{K \times 768}, b_c \in \mathbb{R}^K$ are parameters to learn.
Finally, the action class embedding is normalized as $z = \hat{z} / || \hat{z} ||_2$.

\subsection{Multimodal Video Embedding $\phi(v)$}
\label{sec:video_function}
Unseen action classes may involve both novel spatio-temporal features and objects. For better generalization, we propose to encode videos in two modality streams to capture both the spatio-temporal dynamics and object semantics.

\vspace{0.2em}
\noindent\textbf{Spatio-Temporal Stream in Visual Modality.}
Encouraged by the recent success of 3D CNNs in supervised AR, we employ 3D CNNs, specifically TSM~\cite{lin2019tsm}, to extract Spatio-Temporal (ST) features.
Denote $\bar{x}_v \in \mathbb{R}^{2048}$ as the output from the last pooling layer of TSM, we map $\bar{x}_v$ into the joint embedding space through linear transformation:
\begin{equation}
\hat{x}_v = W_v \bar{x}_v + b_v,
\end{equation}
where $W_v \in \mathbb{R}^{K \times 2048}, b_v \in \mathbb{R}^K$ are parameters to learn.
We also normalize the embedding as $x_v = \hat{x}_v / || \hat{x}_v ||_2$.

\vspace{0.2em}
\noindent\textbf{Object Stream in Text Modality.}
It is a widely acknowledged assumption that the objects associated with actions are prior knowledge available to the ZSAR model~\cite{gao2019know,jain2015objects2action}. We leverage objects automatically recognized from frames to construct a video representation in text modality.
Specifically, we use the BiT model \cite{kolesnikov2019big} pretrained on ImageNet21k dataset to predict object probabilities from evenly sampled 8 frames from video $v$.
The object probabilities over frames are averaged and only the top $N_o$ objects $O=\{o_1, \cdots, o_{N_o}\}$ are preserved and embedded in a concatenated sequence:
\begin{equation}
\label{eqn:video_obj_embed} 
x_{o} = \psi([\text{ED}(o_1); \cdots; \text{ED}(o_{N_o})]),
\end{equation}
where $\text{ED}(o_i)$ denotes ED of object $o_i$ as in Table~\ref{tab:label_defn_examples}. 
Here we use the same $\psi(\cdot)$ function as in the action class embeddings, which explicitly encourages that the object embedding $x_o$ from videos and action class embedding $z$ from action names to lie in the same semantic space.

\vspace{0.2em}
\noindent\textbf{Multimodal Channel Attention.}
The awareness of object semantics can focus the spatio-temporal stream to object-highlighted video features, while object semantics can be enriched with motion features.
We thus further propose to dynamically fuse the two embeddings, $x_v$ and $x_o$, to enhance each other. The formula of injecting $x_o$ to improve $x_v$ is as follows:
\begin{eqnarray}
& g_{vo} = \sigma(W^{2}_{vo}\text{RELU}(W^{1}_{vo} [x_v; x_o])), \\
& x_{vo} = x_{v} g_{vo} / ||x_{v} g_{vo}||_2,
\end{eqnarray}
where $W^{1}_{vo} \in \mathbb{R}^{2K \times K}, W^{2}_{vo} \in \mathbb{R}^{K \times K}$ are parameters, $\sigma$ is the sigmoid function.
Similarly, we obtain $x_{ov}$ from object embedding $x_o$ with guidance of $x_v$. 
Therefore, our video encoder $\phi(v)$ produces two video embeddings $x_{vo}$ and $x_{ov}$ to comprehensively represent the videos.

\subsection{Elaborative Rehearsal (ER) enhanced Training}
\label{sec:training}
Following the standard ZSAR training, given video $v^n$ of seen classes $\mathcal{S}$, we first generate the video embeddings $[x^n_{vo}, x^n_{ov}] = \phi(v^n)$ and action class embedding matrix $Z \in \mathbb{R}^{K \times S}$ where each column $z_i = \psi(i)$. Then, we compute video-action similarity scores:
\begin{equation}
p^{n}_{v} = x_{vo} \cdot Z, \quad p^{n}_{o} = x_{ov} \cdot Z,
\end{equation}
where $\cdot$ denote vector-matrix multiplication, $p^n_{v}, p^{n}_{o} \in \mathbb{R}^{S}$. 
As the negative score between object and action class embeddings mainly indicates that the recognized objects are irrelevant to the action, the magnitude is less important. We thus can fuse the two similarity scores as: 
\begin{equation}
    \label{eqn:vo_prob_fusion}
	p^n = p^n_{v} + \text{max}(p^n_{o}, 0).
\end{equation}

We use a standard contrastive loss to train the action recognition model. To be generalizable, $p \in \mathbb{R}^{C}$ denotes the predicted score, $q \in \mathbb{R}^{C}$ is the ground-truth label where $q_i = 1$ if the $i$-th label is true otherwise $q_i = 0$, and $C$ is the number of classes. The contrastive loss is:
\begin{equation}
\label{eqn:ct_loss}
	L (p, q)  = \frac{-1}{\sum_{j=1}^{C} q_j} \sum_{i=1}^{C} q_i \text{ log } \frac{\mathrm{exp} (p_i / \tau)}{\sum_{j=1}^{C} \mathrm{exp} (p_j / \tau)},
\end{equation}
where $\tau$ is a temperature hyper-parameter. To conduct action recognition on seen data $\mathcal{D}^s$, we convert label $y^n$ into one-hot vector $q^n$ and the loss is:
\begin{equation}
\label{eqn:act_loss}
L_{ar} = \frac{1}{N} \sum_{n=1}^{N} L(p^n, q^n) + L(p^n_{v}, q^n) + L(p^n_{o}, q^n).
\end{equation}
Because $x_{vo}$ is more powerful than $x_{ov}$, the model trained on $\mathcal{D}^s$ tends to overweight $x_{vo}$ in Eq.~\ref{eqn:vo_prob_fusion} and results in overfitting. Therefore, we average the three losses in Eq.~\ref{eqn:act_loss}.

Moreover, as there are only $S$ semantic labels in $L_{ar}$ as semantic supervision, the learned video and text representations might be less generalizable to more diverse semantics.
To address this problem, we further propose an Elaborative Rehearsal (ER) loss, which rehearses the video representation with semantics from ECs obtained from frame-wise object classification.
Denote $O^n = \{o^n_1, \cdots, o^n_{N_o}\}$  the top recognized objects in video $v^n$, we generate semantic representation $\psi(\text{ED}(o^n_i))$ for each $o^n_i$.
Since the total number of all objects are large, we sample top few object classes during training for efficiency. Let $Z^o$ be the object class embeddings in a mini-batch of training, and the ER loss is computed as:
\begin{eqnarray}
\label{eqn:er_loss}
L_{er} = \frac{1}{N} \sum_{n=1}^{N} L(p^n_c, q^n_c) + L(p^n_{c, v}, q^n_c) + L(p^n_{c, o}, q^n_c),
\end{eqnarray}
where $p^n_{c, v} = x_{vo} \cdot Z^o, p^n_{c, o} = x_{ov} \cdot Z^o, p^n_c = p^n_{c, v}  + p^n_{c, o}$, and $q^n_c$ is the ground-truth object labels for $v^n$.

Finally, we combine $L_{ar}$ and $L_{er}$ in our ZSAR model training with a balance factor $\lambda$:
\begin{equation}
\label{eqn:combined_loss}
L = L_{ar} + \lambda L_{er}.
\end{equation}
Comparing to Eq.~\ref{eqn:act_loss}, our model trained by Eq.~\ref{eqn:combined_loss} learns a shared $\psi(\cdot)$ from ECs (i.e. $\psi(o_i)$), and ED (i.e., $\psi(y_i)$). The sharing advocates to learn more comprehensive associations between videos and classes in the common semantic space defined by $(\phi(\cdot), \psi(\cdot))$,  and thus leads to better generalization to unseen classes.

In inference, the action class of $v^m \in \mathcal{D}^{t}$ is recognized with the highest similarity score:
\begin{eqnarray}
\hat{y}^m = \argmax_{y \in \mathcal{T}}  (x^{m}_{vo} \cdot \psi(y) + \text{max}(x^m_{ov}\cdot \psi(y), 0))\\
\nonumber \mbox{where} \quad x^{m}_{vo}, x^{m}_{ov} = \phi(v^m).
\end{eqnarray}

\section{Experiments}

\subsection{Datasets and Splits}
\noindent\textbf{Existing ZSAR Benchmarks.}
Olympic Sports~\cite{niebles2010modeling}, HMDB51~\cite{kuehne2011hmdb} and UCF101~\cite{soomro2012ucf101} are the three most popular datasets used in existing ZSAR papers \cite{junior2019zero}, which contain 783, 6766 and 13320 videos of 16, 51, 101 action categories respectively.
For robust evaluation, Xu \etal \cite{xu2017transductive} proposed to evaluate on 50 independent data splits and report the average accuracy and standard deviation.
In each split, videos of 50\% randomly selected classes are used for training and the remaining 50\% classes are held unseen for testing.
We adopt the same data splits as \cite{xu2017transductive} for fair comparison.

There are two major limitations in the above ZSAR protocols.
Firstly, it is problematic to use deep features pre-trained on other large-scale supervised video datasets because there exist overlapped action classes between pre-training classes and testing classes.
Secondly, the size of training and testing data is small which leads to large variations among different data splits, so that abundant numbers of experiments are necessary to evaluate a model.
To address these limitations, Brattoli \etal \cite{brattoli2020rethinking} proposed another setting which excludes classes overlapped with the above testing dataset in pre-training dataset Kinetics.
Nevertheless, their overlapped class selection algorithm is too tender, leaving the testing classes still seen in the training. Moreover, new end-to-end training of video backbones is needed because this setting does not follow the official Kinetics data split.
Therefore, in this work, we propose a more \emph{realistic, convenient and clean ZSAR protocol}.

\vspace{0.2em}
\noindent\textbf{Our Proposed Kinetics ZSAR Benchmark.}
The evolution of the Kinetics dataset \cite{carreira2018short,carreira2017quo} naturally involves increment of new action classes: Kinetics-400 and Kinetics-600 datasets contains 400 and 600 action classes, respectively. Due to some renamed, removed or split classes in Kinetics-600, we obtain 220 new action classes outside of Kinetics-400 after cleaning.
Therefore, we use 400 action classes in Kinetics-400 as seen classes for training. We randomly split the 220 new classes in Kinetics-600 into 60 validation classes and 160 testing classes respectively.
We independently split the classes for three times for robustness evaluation. 
As shown in our experiments, due to the large-size training and testing sets, the variations of different splits are significantly smaller than previous ZSAR benchmarks. In summary, our benchmark contains 212,577 training videos from Kinetics-400 training set, 2,682 validation videos from Kinetics-600 validation set and 14,125 testing videos from Kinetics-600 testing set on average of the three splits.
More details of our evaluation protocol are in the supplementary material.

\subsection{Implementation Details}
For action class embedding, we use a pretrained 12-layer BERT model \cite{devlin2018bert}, and finetune the last two layers if not specified.
For video embedding, we use TSM \cite{lin2019tsm} pretrained on Kinetics-400 in the spatio-temporal stream for Kinetics benchmark, and BiT image model \cite{kolesnikov2019big} pretrained on ImageNet for the other three benchmarks to avoid overlapped action classes in Kinetics; the object stream uses BiT image model~\cite{kolesnikov2019big} pretrained on ImageNet21k \cite{deng2009imagenet} and top-5 objects are selected for each video.
The above backbones are fixed for fast training.
We use one Nvidia TITAN RTX GPU for the experiments.
More details are presented in the supplementary material.
We set the dimensionality $K=512$ of the common semantic space, $\tau=0.1, \lambda=1$ in the loss and use top-5 objects in the ER loss.
We use ADAM algorithm to train the model with weight decay of 1e-4. The base learning rate is 1e-4  with warm-up and cosine annealing. The model was trained for 10 epochs except on the Olympic Sports dataset where we train 100 epochs due to its small training size. 
The best epoch is selected according to performance on the validation set.
Top-1 and top-5 accuracies (\%) are used to evaluate all models.

\begin{table}
	\centering
	\footnotesize
	\tabcolsep=0.1cm
	\begin{tabular}{lccccccc} \toprule
		Method & Video & Class & Olympics & HMDB51 & UCF101 \\ \midrule
		DAP \cite{lampert2009learning}   & FV & A & 45.4 $\pm$ 12.8 & N/A & 15.9 $\pm$ 1.2 \\
		IAP  \cite{lampert2009learning}  & FV & A & 42.3 $\pm$ 12.5 & N/A & 16.7 $\pm$ 1.1 \\
		HAA \cite{liu2011recognizing}  & FV & A & 46.1 $\pm$ 12.4 & N/A & 14.9 $\pm$ 0.8 \\
		SVE \cite{xu2015semantic}  & BoW & W$_N$ & N/A & 13.0 $\pm$ 2.7 & 10.9 $\pm$ 1.5 \\
		ESZSL  \cite{romera2015embarrassingly} & FV & W$_N$ & 39.6 $\pm$ 9.6 & 18.5 $\pm$ 2.0 & 15.0 $\pm$ 1.3 \\
		SJE \cite{akata2015evaluation}  & FV & W$_N$ & 28.6 $\pm$ 4.9 & 13.3 $\pm$ 2.4 & 9.9 $\pm$ 1.4 \\
		SJE  \cite{akata2015evaluation} & FV & A & 47.5 $\pm$ 14.8 & N/A & 12.0 $\pm$ 1.2 \\
		MTE  \cite{xu2016multi} & FV & W$_N$ & 44.3 $\pm$ 8.1 & 19.7 $\pm$ 1.6 & 15.8 $\pm$ 1.3 \\
		ZSECOC \cite{qin2017zero} & FV & W$_N$ & 59.8 $\pm$ 5.6 & 22.6 $\pm$ 1.2 & 15.1 $\pm$ 1.7 \\
		UR \cite{zhu2018towards} & FV & W$_N$ & N/A & 24.4 $\pm$ 1.6 & 17.5 $\pm$ 1.6 \\
		O2A \cite{jain2015objects2action}  & Obj$^\dag$ & W$_N$ & N/A & 15.6 & 30.3 \\
		ASR \cite{wang2017alternative} & C3D$^*$ & W$_T$ & N/A & 21.8 $\pm$ 0.9 & 24.4 $\pm$ 1.0 \\
		TS-GCN \cite{gao2019know} & Obj$^\dag$ & W$_N$ & 56.5 $\pm$ 6.6 & 23.2 $\pm$ 3.0 & 34.2 $\pm$ 3.1 \\
		E2E \cite{brattoli2020rethinking} & r(2+1)d$^*$ & W$_N$ & N/A & 32.7 & 48 \\ \midrule
		Ours  & (S+Obj)$^\dag$ & ED & \textbf{60.2 $\pm$ 8.9} & \textbf{35.3 $\pm$ 4.6} & \textbf{51.8 $\pm$ 2.9}  \\
	\bottomrule
	\end{tabular}
	\caption{ZSAR performances on the three existing benchmarks. Video: fisher vector (FV), bag of words (BoW), object (Obj), image spatial feature (S), $^*$(trained on video datasets), $^\dag$(trained on ImageNet dataset); Class: attribute (A), word embedding of class names (W$_N$), word embedding of class texts (W$_T$), elaborative description (ED). The average top-1 accuracy (\%) $\pm$ standard deviation are reported.}
	\label{tab:three_datasets_results}
\end{table}

\subsection{Evaluation on Existing ZSAR Benchmarks}
We compare our model with:
(1) Direct/Indirect Attribute Prediction (DAP, IAP) \cite{lampert2009learning};
(2) Human Actions by Attribute (HAA) \cite{liu2011recognizing};
(3) Self-training method with SVM and semantic Embedding (SVE) \cite{xu2015semantic};
(4) Embarrassingly Simple Zero-Shot Learning (ESZSL) \cite{romera2015embarrassingly};
(5) Structured Joint Embedding (SJE) \cite{akata2015evaluation};
(6) Multi-Task Embedding (MTE) \cite{xu2016multi};
(7) Zero-Shot with Error-Correcting Output Codes (ZSECOC) \cite{qin2017zero};
(8) Universal Representation (UR) model \cite{zhu2018towards};
(9) Objects2Action (O2A) \cite{jain2015objects2action};
(10) Alternative Semantic Representation (ASR) \cite{wang2017alternative}, which uses text descriptions and images as alternative class embedding;
(11) TS-GCN \cite{gao2019know} which builds graphs among action and object classes with ConceptNet for better action class embedding;
(12) End-to-End Training (E2E) \cite{brattoli2020rethinking} which uses a reduced Kinetics training set by excluding part of action classes overlapped with testset.
All above methods are evaluated on the inductive ZSL setting, where the videos of unseen action classes are unavailable during training. The unseen action classes are not used in training except \cite{gao2019know}.

Table~\ref{tab:three_datasets_results} presents the comparison.
To avoid leaking information from features pretrained on Kinetics video dataset, we only use image features and predicted objects from a 2D network pretrained on ImageNet \cite{kolesnikov2019big} for video semantic representation learning.
The proposed ER-enhanced ZSAR model achieves consistent improvements over state-of-the-art approaches on three benchmarks.
Our model outperforms previous best performances (without using pretrained video features) with 0.4, 10.9 and 17.6 absolute gains on OlympicSports16, HMDB51 and UCF101 respectively, and achieves even better performance than E2E trained on large-scale Kinetics dataset with 2.6 and 3.8 gains on HMDB51 and UCF101 datasets\footnote{We observe large performance variations with different random weight initialization, which mainly results from the small training set.}.
This demonstrates the effectiveness of our ED as action semantic representation and the ER objective to improve generalization ability of the model.

\begin{table}
	\centering
	\small
	\begin{tabular}{lcccc} \toprule
		Method & Video & Class & top-1 & top-5 \\ \midrule
		DEVISE \cite{frome2013devise} & \multirow{6}{*}{ST} & \multirow{6}{*}{W$_N$} & 23.8 $\pm$ 0.3 & 51.0 $\pm$ 0.6 \\
		ALE \cite{akata2015label} &  &  & 23.4 $\pm$ 0.8 & 50.3 $\pm$ 1.4 \\
		SJE \cite{akata2015evaluation} &  &  & 22.3 $\pm$ 0.6 & 48.2 $\pm$ 0.4 \\
		DEM \cite{zhang2017learning} &  &  & 23.6 $\pm$ 0.7 & 49.5 $\pm$ 0.4 \\
		ESZSL \cite{romera2015embarrassingly} &  &  & 22.9 $\pm$ 1.2 & 48.3 $\pm$ 0.8 \\
		GCN \cite{ghosh2020all} &  &  & 22.3 $\pm$ 0.6 & 49.7 $\pm$ 0.6 \\  \midrule
		\multirow{2}{*}{Ours} & ST & \multirow{2}{*}{ED} & 37.1 $\pm$ 1.7 & 69.3 $\pm$ 0.8 \\
		& ST+Obj &  & \textbf{42.1 $\pm$ 1.4} & \textbf{73.1 $\pm$ 0.3} \\ \bottomrule
	\end{tabular}
	\caption{ZSAR performance on the proposed Kinetics benchmark. Notations are the same as Table~\ref{tab:three_datasets_results}; ST: spatio-temporal feature.}
	\label{tab:kinetics_sota_cmpr}
\end{table}

\subsection{Evaluation on Kinetics ZSAR Benchmark}
Due to limitations of existing benchmarks, we further carry out extensive experiments on more realistic Kinetics ZSAR setting to evaluate the effectiveness of our model.

\subsubsection{Comparison with State of the Arts}
We re-implement state-of-the-art ZSL algorithms on the proposed benchmark, including:
(1) DEVISE \cite{frome2013devise}; (2) ALE \cite{akata2015label}; (3) SJE \cite{akata2015evaluation}; (4) DEM \cite{zhang2017learning}; (5) ESZSL \cite{romera2015embarrassingly}; and (6) GCN \cite{ghosh2020all}: a very recent ZSAR work leveraging knowledge graphs of action classes to predict classification weights as \cite{wang2018zero}.
Details are in the supplementary material.  

Table~\ref{tab:kinetics_sota_cmpr} shows the ZSAR performances of above methods.
When using the same Spatio-Temporal(ST) features extracted from TSM network, our ER-enhanced model with ED and ER loss significantly outperforms previous works with 13.3 and 18.3 absolute gains on top-1 and top-5 accuracies respectively.
The existing methods however achieves similar performances, which might due to ambiguous word embedding representations.
After fusing with object semantics in video semantic representation, the performance of our model gets another boost, demonstrating that ST visual features and object textual features are complementary.
Moreover, compared with the results in Table~\ref{tab:three_datasets_results}, the performance variations on different splits are much lower than those in previous benchmarks, which further proves the superiority of our benchmark for future ZSAR research.

\subsubsection{Ablation Studies}
\begin{table*}[ht]
    \small
    \begin{subtable}[h]{0.23\linewidth}
        \centering
        \begin{tabular}{c|cc} \toprule
        Class Rep & top-1 & top-5 \\ \midrule
        W$_N$ & 26.5 $\pm$ 0.4 & 54.7 $\pm$ 1.2 \\
        Wiki & 25.8 $\pm$ 1.1 & 50.4 $\pm$ 1.6 \\
        Dict & 22.3 $\pm$ 0.4 & 49.7 $\pm$ 0.6 \\
        ED & \textbf{31.0 $\pm$ 1.2}  & \textbf{63.2 $\pm$ 0.4} \\ \bottomrule
       \end{tabular}
       \caption{Comparing action class texts.}
       \label{tab:action_class_text_ablation}
    \end{subtable}
    \hfill
    \begin{subtable}[h]{0.24\linewidth}
        \centering
        \begin{tabular}{c|cc} \toprule
        Class Enc & top-1 & top-5 \\ \midrule
        AvgPool & 25.3 $\pm$ 1.2 & 54.7 $\pm$ 0.6 \\
        AttnPool & 28.2 $\pm$ 1.0 & 56.9 $\pm$ 0.2 \\
        RNN & 25.4 $\pm$ 1.1 & 53.7 $\pm$ 1.1 \\
        BERT & \textbf{31.0 $\pm$ 1.2} & \textbf{63.2 $\pm$ 0.4} \\ \bottomrule
        \end{tabular}
        \caption{Comparing action class encoders.}
        \label{tab:action_class_encoder_ablation}
    \end{subtable}
    \hfill
    \begin{subtable}[h]{0.33\linewidth}
        \centering
        \begin{tabular}{cc|cc} \toprule
        Video & ER & top-1 & top-5 \\ \midrule
        ST & w/o & 31.0 $\pm$ 1.2 & 63.2 $\pm$ 0.4 \\
        ST & w/ & 37.1 $\pm$ 1.7 & 69.3 $\pm$ 0.8 \\
        Obj & w/o & 34.6 $\pm$ 1.4 & 60.6 $\pm$ 1.1 \\
        Obj & w/ & 36.7 $\pm$ 1.0 & 63.2 $\pm$ 0.5 \\ \bottomrule
        \end{tabular}
        \caption{Comparing models with or without ER loss.}
        \label{tab:er_loss_ablation}
    \end{subtable}
    \hfill
    \hfill
    \begin{subtable}[h]{0.4\linewidth}
        \centering
        \begin{tabular}{c|cc} \toprule
        Video & top-1 & top-5 \\ \midrule
        ST & 37.1 $\pm$ 1.7 & 69.3 $\pm$ 0.8 \\
        Obj & 36.7 $\pm$ 1.0 & 63.2 $\pm$ 0.5 \\
        ST + Obj & \textbf{42.1 $\pm$ 1.4} & \textbf{73.1 $\pm$ 0.3} \\ \bottomrule
        \end{tabular}
        \caption{Comparing video representations.}
        \label{tab:video_st_ablation}
    \end{subtable}
    \hfill
    \begin{subtable}[h]{0.55\linewidth}
        \centering
    	\begin{tabular}{c|c|cc} \toprule
    		Video & Loss & top-1 & top-5 \\ \midrule
    		Obj (Name) & ER (Name) & 34.5 $\pm$ 1.6 & 61.4 $\pm$ 1.2\\
    		Obj (ED) & ER (Name) & 36.3 $\pm$ 1.3 & 62.8 $\pm$	0.9  \\
    		Obj (ED) & ER (ED) & 36.7 $\pm$ 1.0 & 63.2 $\pm$ 0.5 \\ \bottomrule
    	\end{tabular}
    	\caption{Comparing EDs and class names to represent object classes.}
    	\label{tab:ed_obj_ablation}
    \end{subtable}
    \caption{Ablation studies on the Kinetics ZSAR benchmark.}
    \label{tab:kinetics_ablation_results}
\end{table*}


We present the following \emph{Q\&As} to prove the effectiveness of our proposed semantic representations and the ER training objectives. More hyper-parameter ablation and analysis are in the supplementary material.
All the ablation studies below are carried out on the Kinetics ZSAR benchmark.

\textbf{Is human involvement necessary for action class representation?}
In Table~\ref{tab:action_class_text_ablation}, we compare different action class representations including action class names (W$_N$), Wikipedia entries (Wiki), Dictionary definitions (Dict) and the manually modified EDs.
All the models use TSM video features and the $L_{ar}$ objective in Eq.~\ref{eqn:act_loss} for training. The W$_N$ is encoded with pre-trained Glove word embedding while others are encoded by BERT because we observe that BERT is not suitable to encode short text such as the class names.
We can see that the automatically crawled texts of the action class are very noisy which are even inferior to the ambiguous class names.
However, with a minimal manual clean of crawled descriptions, we achieve significant improvements such as 8.5\% absolute gains on top-5 accuracy compared to W$_N$. 
This proves that even such easy human involvement is beneficial to the class representation quality as justified in Section~\ref{sec:elaborative_description}, and ED is more discriminative action class prototype than word embedding.

\textbf{How much improvements are from the pre-trained BERT model?}
In Table~\ref{tab:action_class_encoder_ablation}, we compare different action class encoding modules for EDs.
AvgPool, AttnPool and RNN all transfer knowledge from a pre-trained Glove word embedding, and apply average pooling, attentive weighted pooling and bi-directional GRU respectively to encode the ED sentence.
Similar to Table~\ref{tab:action_class_text_ablation}, all the models use TSM video features and are trained with $L_{ar}$.
The pretrained BERT significantly boosts the performance over the other three encoding modules, demonstrating its effectiveness to understand action descriptions.

\textbf{Is the ER loss beneficial?}
Table~\ref{tab:er_loss_ablation} compares models trained with or without ER loss.
The generalization ability on unseen actions is boosted by a large margin through the ER-enhanced training for both ST and object features.
The ER loss augments the semantic labels for videos from automatic elaborative concepts, making the features more generalizable to unseen classes.

\textbf{Whether ST features and object features are complementary?}
The object features alone ``Obj" in Table~\ref{tab:video_st_ablation} are comparable with ST features on top-1 accuracy, but are worse than ST on top-5 accuracy. Their combination ``ST+Obj" via the proposed multimodal channel attention achieves the best performance on the Kinetics ZASR setting.
This shows that object features alone are not discriminative enough, compared to ST features, to differentiate actions. But adding object features enriches ST with the shared semantic embedding among the action classes.

\textbf{Whether EDs are universal representations for both actions and objects?}
Though we show that ED is beneficial to represent action classes, it remains a question whether ED also improves semantic representation for objects.
To be noted, the EDs for objects are automatically extracted from WordNet thanks to the good correspondence between ImageNet classes and WordNet concepts.
Therefore, we replace the ED with the class name of the object in Eq.~\ref{eqn:video_obj_embed} for video object embedding, and in Eq.~\ref{eqn:er_loss} for the ER training objective.
From Table~\ref{tab:ed_obj_ablation}, we see that even though objects are less ambiguous than actions, it is still beneficial to use its ED instead of its class name.

We provide more hyper-parameter ablations and analysis in the supplementary material.

\subsection{Comparison with Supervised Learning}
Previous ZSAR works mainly benchmark the progress with respect to zero-shot methods. However, it is interesting to know how well the state-of-the-arts ZSAR methods really work from a practical prospective of video action recognition. We present one of the first attempts for this purpose. 

In Table~\ref{tab:few_shot_comparison}, we compare our ZSAR model with supervised models trained with different numbers of labeled videos of unseen classes in our Kinetics ZSAR benchmark.
To avoid overfitting on few training samples, we use the same fixed ST features from TSM and only train a linear classification layer. 
It servers as a simple but strong baseline for few-shot learning as suggested in \cite{tian2020rethinking}.
Our ER-enhanced ZSAR model improves over the 1-shot baseline by a large margin, but is still inferior to the model using 2 labeled videos per classes. 
Although our work is the new state-of-the-arts in Table~\ref{tab:three_datasets_results} and \ref{tab:kinetics_sota_cmpr}, it only establishes a starting point from which ZSAR models are comparable with supervised models trained on very few samples.


\begin{table}
	\centering
	\small
	\begin{tabular}{cccc} \toprule
		& \#videos per class & Top-1 (\%) & Top-5 (\%) \\ \midrule
		ER-ZSAR & 0 &42.1 $\pm$ 1.4 & 73.1 $\pm$ 0.3  \\ \midrule
		\multirow{4}{*}{supervised} & 1 & 31.8 $\pm$ 0.8 & 60.2 $\pm$ 2.5 \\
		& 2 & 45.0 $\pm$ 0.9 & 73.2 $\pm$ 0.6 \\
		& 5 & 56.5 $\pm$ 1.5 & 83.4 $\pm$ 0.8 \\
		& 100 & 72.7 $\pm$ 1.4 & 93.3 $\pm$ 0.5 \\ \bottomrule
	\end{tabular}
	\caption{Comparison of our ER-enhanced ZSAR model and supervised few-shot baselines on the Kinetics benchmark.}
	\label{tab:few_shot_comparison}
	\vspace{-0.2cm}
\end{table}

\section{Conclusion}
We present an Elaborative Rehearsal (ER) enhanced model to advance video understanding under the zero-shot setting. 
Our ER-enhanced ZSAR model leverages Elaborative Descriptions (EDs) to learn discriminative semantic representation for action classes, and generates Elaborative Concepts (ECs) from prior knowledge of image-based classification to learn generalizable video semantic representations.
Our model achieves state-of-the-art performances on existing ZSAR benchmarks as well as our newly proposed more realistic ZSAR setting based on the Kinetics dataset.
We demonstrated the potential that our new state-of-the-art on ZSAR benchmarks start to catch up with the supervised AR baselines.
In the future, we will explore the unification of zero-shot and few-shot for action recognition.



\vspace{0.35em}
{\small \noindent\textbf{Acknowledgments.}
This work was supported by the Intelligence Advanced Research Projects Activity (IARPA) via Department of Interior/ Interior Business Center (DOI/IBC) contract number D17PC00340.}


\begin{thebibliography}{10}\itemsep=-1pt

\bibitem{akata2015label}
Zeynep Akata, Florent Perronnin, Zaid Harchaoui, and Cordelia Schmid.
\newblock Label-embedding for image classification.
\newblock {\em IEEE transactions on pattern analysis and machine intelligence},
  38(7):1425--1438, 2015.

\bibitem{akata2015evaluation}
Zeynep Akata, Scott Reed, Daniel Walter, Honglak Lee, and Bernt Schiele.
\newblock Evaluation of output embeddings for fine-grained image
  classification.
\newblock In {\em Proceedings of the IEEE conference on computer vision and
  pattern recognition}, pages 2927--2936, 2015.

\bibitem{benjamin2000relationship}
Aaron~S Benjamin and Robert~A Bjork.
\newblock On the relationship between recognition speed and accuracy for words
  rehearsed via rote versus elaborative rehearsal.
\newblock {\em Journal of Experimental Psychology: Learning, Memory, and
  Cognition}, 26(3):638, 2000.

\bibitem{brattoli2020rethinking}
Biagio Brattoli, Joseph Tighe, Fedor Zhdanov, Pietro Perona, and Krzysztof
  Chalupka.
\newblock Rethinking zero-shot video classification: End-to-end training for
  realistic applications.
\newblock In {\em Proceedings of the IEEE/CVF Conference on Computer Vision and
  Pattern Recognition}, pages 4613--4623, 2020.

\bibitem{carreira2018short}
Joao Carreira, Eric Noland, Andras Banki-Horvath, Chloe Hillier, and Andrew
  Zisserman.
\newblock A short note about kinetics-600.
\newblock {\em arXiv preprint arXiv:1808.01340}, 2018.

\bibitem{carreira2017quo}
Joao Carreira and Andrew Zisserman.
\newblock Quo vadis, action recognition? a new model and the kinetics dataset.
\newblock In {\em proceedings of the IEEE Conference on Computer Vision and
  Pattern Recognition}, pages 6299--6308, 2017.

\bibitem{cui2020does}
Leyang Cui, Sijie Cheng, Yu Wu, and Yue Zhang.
\newblock Does bert solve commonsense task via commonsense knowledge?
\newblock {\em arXiv preprint arXiv:2008.03945}, 2020.

\bibitem{deng2009imagenet}
Jia Deng, Wei Dong, Richard Socher, Li-Jia Li, Kai Li, and Li Fei-Fei.
\newblock Imagenet: A large-scale hierarchical image database.
\newblock In {\em 2009 IEEE conference on computer vision and pattern
  recognition}, pages 248--255. Ieee, 2009.

\bibitem{devlin2018bert}
Jacob Devlin, Ming-Wei Chang, Kenton Lee, and Kristina Toutanova.
\newblock Bert: Pre-training of deep bidirectional transformers for language
  understanding.
\newblock {\em arXiv preprint arXiv:1810.04805}, 2018.

\bibitem{feichtenhofer2020x3d}
Christoph Feichtenhofer.
\newblock X3d: Expanding architectures for efficient video recognition.
\newblock In {\em Proceedings of the IEEE/CVF Conference on Computer Vision and
  Pattern Recognition}, pages 203--213, 2020.

\bibitem{feichtenhofer2019slowfast}
Christoph Feichtenhofer, Haoqi Fan, Jitendra Malik, and Kaiming He.
\newblock Slowfast networks for video recognition.
\newblock In {\em Proceedings of the IEEE international conference on computer
  vision}, pages 6202--6211, 2019.

\bibitem{frome2013devise}
Andrea Frome, Greg~S Corrado, Jon Shlens, Samy Bengio, Jeff Dean, Marc'Aurelio
  Ranzato, and Tomas Mikolov.
\newblock Devise: A deep visual-semantic embedding model.
\newblock In {\em Advances in neural information processing systems}, pages
  2121--2129, 2013.

\bibitem{gan2016concepts}
Chuang Gan, Ming Lin, Yi Yang, Gerard De~Melo, and Alexander~G Hauptmann.
\newblock Concepts not alone: Exploring pairwise relationships for zero-shot
  video activity recognition.
\newblock In {\em Thirtieth AAAI conference on artificial intelligence}, 2016.

\bibitem{gan2016learning}
Chuang Gan, Tianbao Yang, and Boqing Gong.
\newblock Learning attributes equals multi-source domain generalization.
\newblock In {\em Proceedings of the IEEE conference on computer vision and
  pattern recognition}, pages 87--97, 2016.

\bibitem{gao2019know}
Junyu Gao, Tianzhu Zhang, and Changsheng Xu.
\newblock I know the relationships: Zero-shot action recognition via two-stream
  graph convolutional networks and knowledge graphs.
\newblock In {\em Proceedings of the AAAI Conference on Artificial
  Intelligence}, volume~33, pages 8303--8311, 2019.

\bibitem{ghadiyaram2019large}
Deepti Ghadiyaram, Du Tran, and Dhruv Mahajan.
\newblock Large-scale weakly-supervised pre-training for video action
  recognition.
\newblock In {\em Proceedings of the IEEE Conference on Computer Vision and
  Pattern Recognition}, pages 12046--12055, 2019.

\bibitem{ghosh2020all}
Pallabi Ghosh, Nirat Saini, Larry~S Davis, and Abhinav Shrivastava.
\newblock All about knowledge graphs for actions.
\newblock {\em arXiv preprint arXiv:2008.12432}, 2020.

\bibitem{goodfellow2016deep}
Ian Goodfellow, Yoshua Bengio, and Aaron Courville.
\newblock {\em Deep learning}.
\newblock MIT press, 2016.

\bibitem{he2016deep}
Kaiming He, Xiangyu Zhang, Shaoqing Ren, and Jian Sun.
\newblock Deep residual learning for image recognition.
\newblock In {\em Proceedings of the IEEE conference on computer vision and
  pattern recognition}, pages 770--778, 2016.

\bibitem{huynh2020fine}
Dat Huynh and Ehsan Elhamifar.
\newblock Fine-grained generalized zero-shot learning via dense attribute-based
  attention.
\newblock In {\em Proceedings of the IEEE/CVF Conference on Computer Vision and
  Pattern Recognition}, pages 4483--4493, 2020.

\bibitem{jain2015objects2action}
Mihir Jain, Jan~C van Gemert, Thomas Mensink, and Cees~GM Snoek.
\newblock Objects2action: Classifying and localizing actions without any video
  example.
\newblock In {\em Proceedings of the IEEE international conference on computer
  vision}, pages 4588--4596, 2015.

\bibitem{junior2019zero}
Valter Lu{\'\i}s~Estevam Junior, Helio Pedrini, and David Menotti.
\newblock Zero-shot action recognition in videos: A survey.
\newblock {\em arXiv preprint arXiv:1909.06423}, 2019.

\bibitem{karpathy2014large}
Andrej Karpathy, George Toderici, Sanketh Shetty, Thomas Leung, Rahul
  Sukthankar, and Li Fei-Fei.
\newblock Large-scale video classification with convolutional neural networks.
\newblock In {\em Proceedings of the IEEE conference on Computer Vision and
  Pattern Recognition}, pages 1725--1732, 2014.

\bibitem{kolesnikov2019big}
Alexander Kolesnikov, Lucas Beyer, Xiaohua Zhai, Joan Puigcerver, Jessica Yung,
  Sylvain Gelly, and Neil Houlsby.
\newblock Big transfer (bit): General visual representation learning.
\newblock {\em arXiv preprint arXiv:1912.11370}, 2019.

\bibitem{kuehne2011hmdb}
Hildegard Kuehne, Hueihan Jhuang, Est{\'\i}baliz Garrote, Tomaso Poggio, and
  Thomas Serre.
\newblock Hmdb: a large video database for human motion recognition.
\newblock In {\em 2011 International Conference on Computer Vision}, pages
  2556--2563. IEEE, 2011.

\bibitem{lampert2009learning}
Christoph~H Lampert, Hannes Nickisch, and Stefan Harmeling.
\newblock Learning to detect unseen object classes by between-class attribute
  transfer.
\newblock In {\em 2009 IEEE Conference on Computer Vision and Pattern
  Recognition}, pages 951--958. IEEE, 2009.

\bibitem{li2017zero}
Yanan Li, Donghui Wang, Huanhang Hu, Yuetan Lin, and Yueting Zhuang.
\newblock Zero-shot recognition using dual visual-semantic mapping paths.
\newblock In {\em Proceedings of the IEEE Conference on Computer Vision and
  Pattern Recognition}, pages 3279--3287, 2017.

\bibitem{lin2019tsm}
Ji Lin, Chuang Gan, and Song Han.
\newblock Tsm: Temporal shift module for efficient video understanding.
\newblock In {\em Proceedings of the IEEE International Conference on Computer
  Vision}, pages 7083--7093, 2019.

\bibitem{liu2011recognizing}
Jingen Liu, Benjamin Kuipers, and Silvio Savarese.
\newblock Recognizing human actions by attributes.
\newblock In {\em CVPR 2011}, pages 3337--3344. IEEE, 2011.

\bibitem{mandal2019out}
Devraj Mandal, Sanath Narayan, Sai~Kumar Dwivedi, Vikram Gupta, Shuaib Ahmed,
  Fahad~Shahbaz Khan, and Ling Shao.
\newblock Out-of-distribution detection for generalized zero-shot action
  recognition.
\newblock In {\em Proceedings of the IEEE Conference on Computer Vision and
  Pattern Recognition}, pages 9985--9993, 2019.

\bibitem{miller1995wordnet}
George~A Miller.
\newblock Wordnet: a lexical database for english.
\newblock {\em Communications of the ACM}, 38(11):39--41, 1995.

\bibitem{niebles2010modeling}
Juan~Carlos Niebles, Chih-Wei Chen, and Li Fei-Fei.
\newblock Modeling temporal structure of decomposable motion segments for
  activity classification.
\newblock In {\em European conference on computer vision}, pages 392--405.
  Springer, 2010.

\bibitem{paz2020zest}
Tzuf Paz-Argaman, Yuval Atzmon, Gal Chechik, and Reut Tsarfaty.
\newblock Zest: Zero-shot learning from text descriptions using textual
  similarity and visual summarization.
\newblock {\em arXiv preprint arXiv:2010.03276}, 2020.

\bibitem{qiao2016less}
Ruizhi Qiao, Lingqiao Liu, Chunhua Shen, and Anton Van Den~Hengel.
\newblock Less is more: zero-shot learning from online textual documents with
  noise suppression.
\newblock In {\em Proceedings of the IEEE Conference on Computer Vision and
  Pattern Recognition}, pages 2249--2257, 2016.

\bibitem{qin2017zero}
Jie Qin, Li Liu, Ling Shao, Fumin Shen, Bingbing Ni, Jiaxin Chen, and Yunhong
  Wang.
\newblock Zero-shot action recognition with error-correcting output codes.
\newblock In {\em Proceedings of the IEEE Conference on Computer Vision and
  Pattern Recognition}, pages 2833--2842, 2017.

\bibitem{qiu2017learning}
Zhaofan Qiu, Ting Yao, and Tao Mei.
\newblock Learning spatio-temporal representation with pseudo-3d residual
  networks.
\newblock In {\em proceedings of the IEEE International Conference on Computer
  Vision}, pages 5533--5541, 2017.

\bibitem{romera2015embarrassingly}
Bernardino Romera-Paredes and Philip Torr.
\newblock An embarrassingly simple approach to zero-shot learning.
\newblock In {\em International Conference on Machine Learning}, pages
  2152--2161, 2015.

\bibitem{simonyan2014two}
Karen Simonyan and Andrew Zisserman.
\newblock Two-stream convolutional networks for action recognition in videos.
\newblock In {\em Advances in neural information processing systems}, pages
  568--576, 2014.

\bibitem{soomro2012ucf101}
Khurram Soomro, Amir~Roshan Zamir, and Mubarak Shah.
\newblock Ucf101: A dataset of 101 human actions classes from videos in the
  wild.
\newblock {\em arXiv preprint arXiv:1212.0402}, 2012.

\bibitem{tian2020rethinking}
Yonglong Tian, Yue Wang, Dilip Krishnan, Joshua~B Tenenbaum, and Phillip Isola.
\newblock Rethinking few-shot image classification: a good embedding is all you
  need?
\newblock In {\em Computer Vision--ECCV 2020: 16th European Conference,
  Glasgow, UK, August 23--28, 2020, Proceedings, Part XIV 16}, pages 266--282.
  Springer, 2020.

\bibitem{tran2015learning}
Du Tran, Lubomir Bourdev, Rob Fergus, Lorenzo Torresani, and Manohar Paluri.
\newblock Learning spatiotemporal features with 3d convolutional networks.
\newblock In {\em Proceedings of the IEEE international conference on computer
  vision}, pages 4489--4497, 2015.

\bibitem{tran2018closer}
Du Tran, Heng Wang, Lorenzo Torresani, Jamie Ray, Yann LeCun, and Manohar
  Paluri.
\newblock A closer look at spatiotemporal convolutions for action recognition.
\newblock In {\em Proceedings of the IEEE conference on Computer Vision and
  Pattern Recognition}, pages 6450--6459, 2018.

\bibitem{wang2013action}
Heng Wang and Cordelia Schmid.
\newblock Action recognition with improved trajectories.
\newblock In {\em Proceedings of the IEEE international conference on computer
  vision}, pages 3551--3558, 2013.

\bibitem{wang2018temporal}
Limin Wang, Yuanjun Xiong, Zhe Wang, Yu Qiao, Dahua Lin, Xiaoou Tang, and Luc
  Van~Gool.
\newblock Temporal segment networks for action recognition in videos.
\newblock {\em IEEE transactions on pattern analysis and machine intelligence},
  41(11):2740--2755, 2018.

\bibitem{wang2017alternative}
Qian Wang and Ke Chen.
\newblock Alternative semantic representations for zero-shot human action
  recognition.
\newblock In {\em Joint European Conference on Machine Learning and Knowledge
  Discovery in Databases}, pages 87--102. Springer, 2017.

\bibitem{wang2018non}
Xiaolong Wang, Ross Girshick, Abhinav Gupta, and Kaiming He.
\newblock Non-local neural networks.
\newblock In {\em Proceedings of the IEEE conference on computer vision and
  pattern recognition}, pages 7794--7803, 2018.

\bibitem{wang2018zero}
Xiaolong Wang, Yufei Ye, and Abhinav Gupta.
\newblock Zero-shot recognition via semantic embeddings and knowledge graphs.
\newblock In {\em Proceedings of the IEEE conference on computer vision and
  pattern recognition}, pages 6857--6866, 2018.

\bibitem{xian2018zero}
Yongqin Xian, Christoph~H Lampert, Bernt Schiele, and Zeynep Akata.
\newblock Zero-shot learning—a comprehensive evaluation of the good, the bad
  and the ugly.
\newblock {\em IEEE transactions on pattern analysis and machine intelligence},
  41(9):2251--2265, 2018.

\bibitem{xu2015semantic}
Xun Xu, Timothy Hospedales, and Shaogang Gong.
\newblock Semantic embedding space for zero-shot action recognition.
\newblock In {\em 2015 IEEE International Conference on Image Processing
  (ICIP)}, pages 63--67. IEEE, 2015.

\bibitem{xu2017transductive}
Xun Xu, Timothy Hospedales, and Shaogang Gong.
\newblock Transductive zero-shot action recognition by word-vector embedding.
\newblock {\em International Journal of Computer Vision}, 123(3):309--333,
  2017.

\bibitem{xu2016multi}
Xun Xu, Timothy~M Hospedales, and Shaogang Gong.
\newblock Multi-task zero-shot action recognition with prioritised data
  augmentation.
\newblock In {\em European Conference on Computer Vision}, pages 343--359.
  Springer, 2016.

\bibitem{zellers2017zero}
Rowan Zellers and Yejin Choi.
\newblock Zero-shot activity recognition with verb attribute induction.
\newblock In {\em Proceedings of the 2017 Conference on Empirical Methods in
  Natural Language Processing}, pages 946--958, 2017.

\bibitem{zhang2017learning}
Li Zhang, Tao Xiang, and Shaogang Gong.
\newblock Learning a deep embedding model for zero-shot learning.
\newblock In {\em Proceedings of the IEEE Conference on Computer Vision and
  Pattern Recognition}, pages 2021--2030, 2017.

\bibitem{zhu2018towards}
Yi Zhu, Yang Long, Yu Guan, Shawn Newsam, and Ling Shao.
\newblock Towards universal representation for unseen action recognition.
\newblock In {\em Proceedings of the IEEE Conference on Computer Vision and
  Pattern Recognition}, pages 9436--9445, 2018.

\end{thebibliography}

\appendix
\section*{Supplementary Material}
\section{Construction of Elaborative Description}
Different from the ImageNet object concepts universally defined in standard dictionaries, there are no standard sources to define action classes. We collect Elaborative Descriptions (ED) for action classes in two steps: firstly automatically crawling candidate sentences to describe action classes from the Internet; then manually selecting or modifying a minimum set of candidate sentences as the EDs.
We release the collected EDs publicly\footnote{ \url{https://github.com/DeLightCMU/ElaborativeRehearsal/blob/main/datasets/Kinetics/zsl220/classes620_label_defn.json}}.


In the first crawling step, we utilize Wikipedia and online dictionaries.
Given an action class such as ``dumpster diving'' as query, we use Wikipedia crawling toolkit\footnote{\url{https://github.com/goldsmith/Wikipedia/}} to collect summary of the first page returned by Wikipedia. This page is usually useful for describing novel actions such as ``photobombing'' and collocations such as ``clean and jerk''. We also let Wikipedia find a relevant page title for the query in case no exact page is matched with the query.
But to be noted, the returned page can be noisy, especially for compositional action classes. For example, the query ``assembling computer'' gets the page ``assembly language'' in computer science.
Therefore, we further crawl dictionary definitions\footnote{\url{https://dictionaryapi.dev/}} for words and phrases in the query action class.
We split crawled data into candidate sentences via spacy toolkit\footnote{\url{https://spacy.io/}}, and remove non-ascii letters in each sentence.

\begin{figure}
	\includegraphics[width=\linewidth]{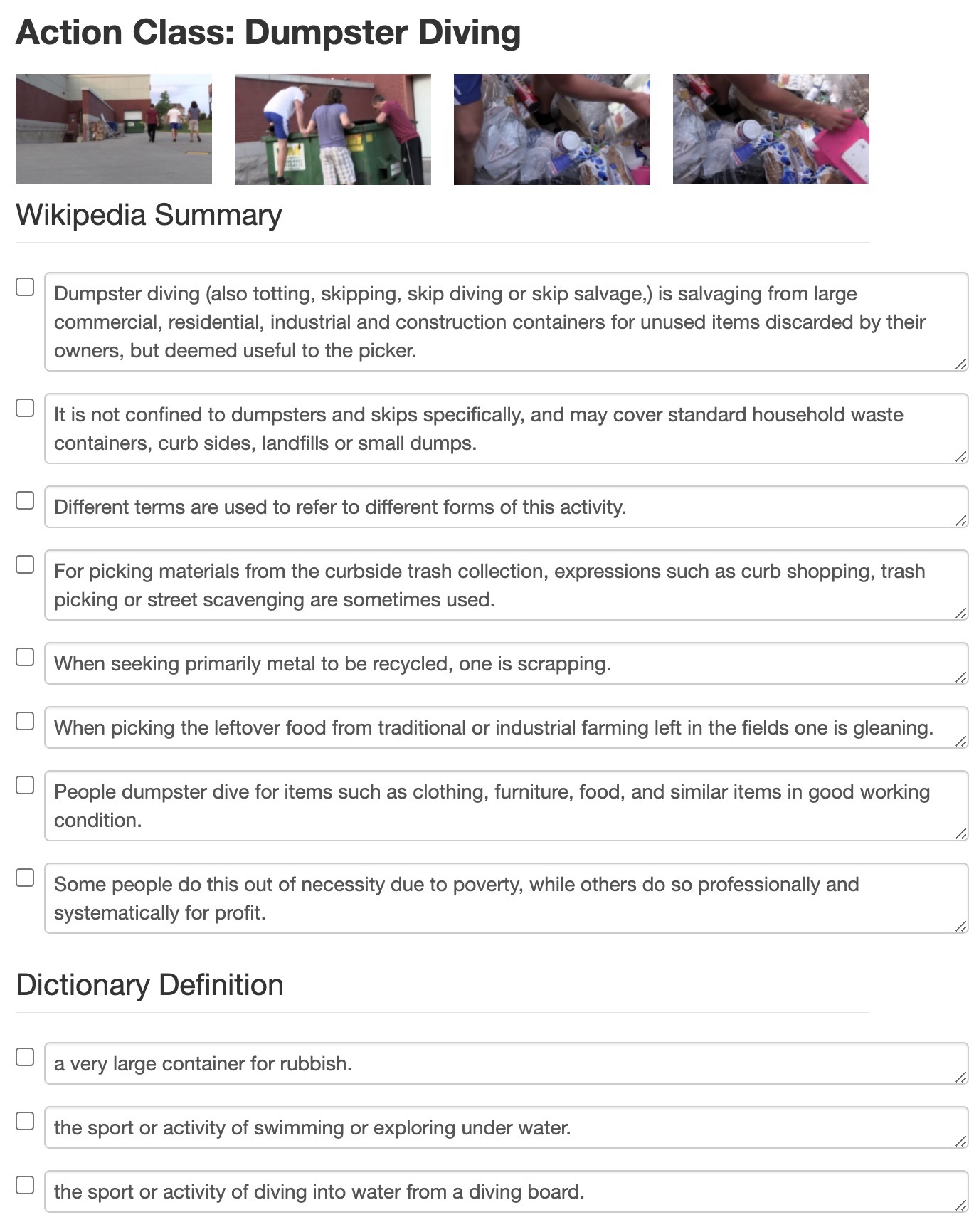}
	\caption{The annotation interface to collect Elaborative Descriptions (ED) for action classes.}
	\label{fig:ED_anno_interface}
\end{figure}

In the second cleaning step, we represent candidate sentences and a video exemplar in a webpage to annotators as shown in Figure~\ref{fig:ED_anno_interface}.
We ask the annotator to select or modify a minimum set of candidate sentences to describe the action class. 
If no candidate sentences are qualified, the annotator can write a new definition.
It takes less than 20s on average to generate the ED per action class. 
The average length of EDs for actions in the Kinetics dataset is 36 words. 

\section{The Proposed Kinetics ZSAR Benchmark}
We use Kinetics-400 \cite{carreira2017quo} as the training dataset and the associated 400 action classes as seen classes. The new classes in Kinetics-600 \cite{carreira2018short}  are used as unseen classes.
Due to some renamed, removed or split classes in the evolution from Kinetics-400 to Kinetics-600, it is problematic to obtain new classes by simply selecting action classes that are not in the original (400) action names set.
In these ambiguous classes, the videos are still the same even if the class names are different in Kinetics-600. Therefore, we further use the overlapping videos as additional cues to find new classes in Kinetics-600.
In this way, we obtained 220 new action classes outside of Kinetics-400.

As mentioned in \cite{xian2018zero}, it is necessary to hold a validation class split that is disjoint from the training and testing classes, to tune hyper-parameters of the zero-shot methods.
Therefore, we randomly split the 220 new classes in Kinetics-600 into the 60 validation and 160 testing classes. 
To avoid the potential bias in only one split, we independently split the classes for three times to improve the robustness of evaluation.
The validation and testing videos are from the original Kinetics-600 splits respectively.
To be noted, since the training set is the same for the three splits, the ZSAR methods only need to train once on the training set, and then different validation sets are used to select the best models for the respective testing sets.
The dataset statistics of the three splits are shown in Table~\ref{tab:kinetics_benchmark}.

\begin{table}
	\centering
	\begin{tabular}{ccccc} \toprule
		\multirow{2}{*}{} & \multirow{2}{*}{\# classes} & \multicolumn{3}{c}{\# videos} \\ 
		&  & split1 & split2 & split3 \\ \midrule
		train & 400 & 212,577 & 212,577 & 212,577 \\
		val & 60 & 2,670 & 2,712 & 2,663 \\
		test & 160 & 14,131 & 14,078 & 14,167 \\ \bottomrule
	\end{tabular}
	\caption{Dataset statistics of our Kinetics ZSAR benchmark.}
	\label{tab:kinetics_benchmark}
\end{table}

\section{Our Implemented Baseline Models}

As there is no baseline to compare on our newly proposed Kinetics ZSAR benchmark, we implement the following state-of-the-art ZSL algorithms:
(1) DEVISE \cite{frome2013devise}; (2) ALE \cite{akata2015label}; (3) SJE \cite{akata2015evaluation}; (4) DEM \cite{zhang2017learning}; (5) ESZSL \cite{romera2015embarrassingly}; and (6) GCN \cite{ghosh2020all}.

Among them, DEVISE, ALE, SJE and ESZSL use bilinear compatibility function to associate video $v$ and class $y$ with different objectives in training:
\begin{equation}
\label{eqn:bilinear_func}
F(v, y; W) = \phi(v)^T W \psi(y)
\end{equation}
All the methods use the same ST video encoding $\phi(v)$ as ours.
The semantic representation $\psi(y)$ for action classes are L2 normalized mean-pooled Glove42b \cite{pennington2014glove} feature of class names, which shows better performance than other word embeddings and sent2vec embeddings \cite{pagliardini2018unsupervised}. 
We revisit the core idea of each method below.

\vspace{0.2em}
\noindent\textbf{\textbf{DEVISE \cite{frome2013devise}}} uses pairwise ranking objective:
\begin{equation}
	\sum_{y \in \mathcal{S}} [\Delta(y^n, y) + F(v^n, y; W) - F(v^n, y^n; W)]_+
\end{equation}
where $\Delta(y^n, y) = 0$ if $y^n = y$ otherwise 0.2.

\vspace{0.2em}
\noindent\textbf{ALE \cite{akata2015label}} uses weighted approximate ranking objective:
\begin{equation}
\sum_{y \in \mathcal{S}}  \frac{l_{r_{\Delta(v^n, y^n)}}}{r_{\Delta(v^n, y^n)}}  [\Delta(y^n, y) + F(v^n, y; W) - F(v^n, y^n; W)]_+
\end{equation}
where $l_k = \sum_{i=1}^k \alpha_i$ and $r_{\Delta(v^n, y^n)}$ is defined as:
\begin{equation}
	\sum_{y \in \mathcal{S}} \mathbf{1}(F(v^n, y; W) + \Delta(y^n, y) \ge F(v^n, y^n; W))
\end{equation}
We use $\alpha_i = 1 / i$ which puts a high emphasis on the top of the rank list.

\vspace{0.2em}
\noindent\textbf{SJE \cite{akata2015evaluation}} uses hard negative label mining with the training objective as follows:
\begin{equation}
\mathrm{max}_{y \in \mathcal{S}} [\Delta(y^n, y) + F(v^n, y; W) - F(v^n, y^n; W)]_+
\end{equation}

\vspace{0.2em}
\noindent\textbf{DEM \cite{zhang2017learning}} uses the visual space as the embedding space, which learns a non-linear mapping from class features to visual features and minimizes the model with MSE loss:
\begin{equation}
	\begin{split}
	\frac{1}{N} \sum_{i=1}^{N} ||\phi(v^n) &- f_1(W_2 f_1(W_1 \psi(y^n))||^2 \\
	&+ \lambda (|| W_1 ||^2 + || W_2 ||^2)
	\end{split}
\end{equation}

\vspace{0.2em}
\noindent\textbf{ESZSL \cite{romera2015embarrassingly}} applies a square loss to the pairwise ranking formulation and adds regularization terms to optimize:
\begin{equation}
	\gamma || W\psi(y) ||^2 + \lambda || \phi(v)^T W ||^2 + \beta || W ||^2
\end{equation}
There exists a closed form solution for the objective.

\vspace{0.2em}
\noindent\textbf{GCN \cite{ghosh2020all}} is a very recent ZSAR work which builds knowledge graphs for action classes to predict classification weights as \cite{wang2018zero}.
We use the first type of knowledge graphs as their work, which is built based on similarity of class embeddings. 
Six GCN layers are used to predict classification weights from the built graph.

\section{More Ablation Studies}

\vspace{0.2em}
\noindent\textbf{Multimodal-based Channel Attention.}
In Table~\ref{tab:mca_expr}, we compare our ER-enhanced models with or without multimodal-based channel attention in video semantic representation encoding in Section 3.3 of our main paper. The comparison shows that the proposed channel attention is beneficial to generate better video semantic representations from the ST and object streams.

\begin{table}
	\centering
	\begin{tabular}{ccc} \toprule
		model & Top-1 (\%) & Top-5 (\%) \\ \midrule
		w/o MCA & 41.0 $\pm$ 1.7 & 71.9 $\pm$ 0.7  \\ 
		w/ MCA & \textbf{42.1 $\pm$ 1.4} & \textbf{73.1 $\pm$ 0.3} \\ \bottomrule
	\end{tabular}
	\caption{Comparison of ER-enhanced models with or without multimodal-based channel attention (MCA) on Kinetics ZSAR benchmark.}
	\label{tab:mca_expr}
\end{table}

\vspace{0.2em}
\noindent\textbf{ER loss.}
Table~\ref{tab:more_er_loss_ablation} presents additional models (using spatial-temporal and object video representations) trained with or without ER loss.
The trend is the same as Table 4c in the main paper. The ER loss improves the generalization ability on unseen actions by 2.6\% on Top-1 accuracy and 3.0\% on Top-5 accuracy.

\begin{table}
\centering
\begin{tabular}{cccc} \toprule
Video & ER & top-1 & top-5 \\ \midrule
\multirow{2}{*}{ST+Obj} & w/o & 39.5 $\pm$ 1.4 & 70.1 $\pm$ 0.6 \\
 & w/ & \textbf{42.1 $\pm$ 1.4} & \textbf{73.1 $\pm$ 0.3} \\ \bottomrule
\end{tabular}
\caption{Comparison of ER-enhanced models with or without ER loss on Kinetics ZSAR benchmark.}
\label{tab:more_er_loss_ablation}
\end{table}

\begin{table}
	\centering
	\begin{tabular}{cccc} \toprule
		\multicolumn{2}{c}{\# objects} & \multirow{2}{*}{Top-1 (\%)} & \multirow{2}{*}{Top-5 (\%)} \\ 
		VE & ER &  &  \\ \midrule
		5 & 0 & 39.5 $\pm$ 1.4 & 70.1 $\pm$ 0.6 \\
		5 & 1 & 41.5 $\pm$ 1.9 & 70.9 $\pm$ 1.0 \\
		5 & 5 & \textbf{42.1 $\pm$ 1.4} & \textbf{73.1 $\pm$ 0.3} \\
		5 & 10 & 41.0 $\pm$ 1.6 & 72.0 $\pm$ 1.2 \\ \midrule
		0 & 5 & 37.1 $\pm$ 1.7 & 69.3 $\pm$ 0.8 \\
		1 & 5 & 37.6 $\pm$ 1.0 & 68.9 $\pm$ 0.8 \\
		5 & 5 & \textbf{42.1 $\pm$ 1.4} & \textbf{73.1 $\pm$ 0.3} \\
		10 & 5 & 42.0 $\pm$ 1.3 & 72.3 $\pm$ 0.6 \\ \bottomrule
	\end{tabular}
	\caption{Comparison of ER-enhanced models using different numbers of objects in object stream of video encoding (VE) and ER loss (ER) on our Kinetics ZSAR benchmark.}
	\label{tab:nobjs_expr}
\end{table}

\vspace{0.2em}
\noindent\textbf{Number of Object Concepts.}
Table~\ref{tab:nobjs_expr} presents ZSAR performances using different numbers of object concepts predicted in the object stream of video encoding and ER loss respectively.
We can see that the ZSAR performance first increases with the number of objects and then decreases, which might result from incorrectly detected (false positive) object concepts.

\begin{table}
	\centering
	\begin{tabular}{ccccc} \toprule
		\multicolumn{2}{c}{Model} & \multirow{2}{*}{ObjSet} & \multirow{2}{*}{Top-1 (\%)} & \multirow{2}{*}{Top-5 (\%)} \\
		Video & Loss &  &  &  \\ \midrule
		ST & AR & - & 31.0 $\pm$ 1.2 & 63.2 $\pm$ 0.4 \\ \midrule
		Obj & AR + ER & 1K & 24.8 $\pm$ 0.7 & 51.7 $\pm$ 0.7 \\
		Obj & AR + ER & 21K & 36.7 $\pm$ 1.0 & 63.2 $\pm$ 0.5 \\ \midrule
		ST + Obj & AR + ER & 1K & 34.7 $\pm$ 1.1 & 67.4 $\pm$ 1.0 \\
		ST + Obj & AR + ER & 21K & 42.1 $\pm$ 1.4 & 73.1 $\pm$ 0.3 \\ \bottomrule
	\end{tabular}
	\caption{Comparison of using different sets of object concepts (``ObjSet") in our ER model on our Kinetics ZSAR benchmark.}
	\label{tab:obj_types_cmpr}
\end{table}

\vspace{0.2em}
\noindent\textbf{Different Object Concepts.}
We compare different sets of object concepts in Table~\ref{tab:nobjs_expr}.
In our main paper, we use the full concept set in ImageNet21k from the BiT model. We compare it with concepts in ImageNet1k from Resnext50\footnote{\url{https://pytorch.org/docs/stable/torchvision/models.html\#classification}} image classification model.
The predicted concepts of the latter are not as accurate as the former due to less training data and fewer concept classes. When only using the object concepts as video semantic representation, we can see that ZSAR performance of the ImageNet1k concepts are much worse than that of ImageNet21k and ST features.
It indicates that the object concepts set and recognition performance are important.
Though objects from ImageNet1k alone are not competitive, they are still complementary to ST video features. The combination of object and ST feature in our full ER model also achieves better performance.

\begin{table}
	\centering
	\begin{tabular}{cccc} \toprule
		\multicolumn{2}{c}{Model} & \multirow{2}{*}{Top-1 (\%)} & \multirow{2}{*}{Top-5 (\%)} \\
		Video & Loss &  &  \\ \midrule
		ST & AR & 31.0 $\pm$ 1.2 & 63.2 $\pm$ 0.4 \\
		ST (NL) & AR + ER & 32.0 $\pm$ 0.9 & 63.9 $\pm$ 0.6 \\ \midrule
		ST + Obj & AR + ER & 42.1 $\pm$ 1.4 & 73.1 $\pm$ 0.3 \\
		ST (NL) + Obj & AR + ER & 42.7 $\pm$ 1.6 & 73.3 $\pm$ 0.6 \\ \bottomrule
	\end{tabular}
	\caption{Comparison of using different Spatio-Temporal (ST) features on Kinetics ZSAR benchmark. \textbf{NL} denotes non-local.}
	\label{tab:st_ft_cmpr}
\end{table}

\vspace{0.2em}
\noindent\textbf{Different Spatio-Temporal(ST) Features.}
We further verify the generalization of our approach on different ST features.
Table~\ref{tab:st_ft_cmpr} shows the results. We compare the TSM model and an enhanced TSM with non-local attentions for ST feature extraction. Better ST features are beneficial to the ZSAR performance.

\begin{figure}
     \centering
     \begin{subfigure}[b]{0.47\columnwidth}
         \centering
         \includegraphics[width=\textwidth]{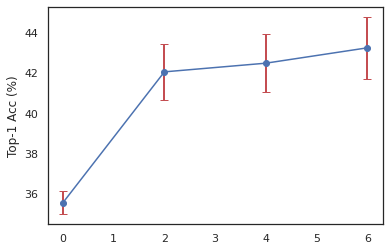}
         \caption{\# of finetuned layers.}
         \label{fig:number_of_bert_layers}
     \end{subfigure}
     \hfill
     \begin{subfigure}[b]{0.47\columnwidth}
         \centering
         \includegraphics[width=\textwidth]{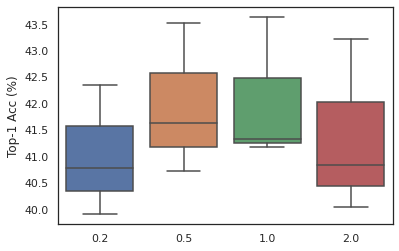}    
         \caption{lambda of ER loss.}
         \label{fig:lambda_ablation}
     \end{subfigure}
    \caption{Top-1 accuracy for different hyper-parameters.}
    \label{fig:hyperparam_ablation_results}
\end{figure}

\vspace{0.2em}
\noindent\textbf{Number of Finetuned Layers in BERT Model.}
As shown in Figure~\ref{fig:number_of_bert_layers}, finetuning more layers in BERT continuously improves the performance, which however consumes more resources, \eg we use 1 RTX 2080Ti to finetune 2 layers in BERT, but need 4 GPUs to finetune 6 layers.

\vspace{0.2em}
\noindent\textbf{$\lambda$ for Elaborative Rehearsal Loss.}
Figure~\ref{fig:lambda_ablation} presents the performance of different $\lambda$s for the ER loss, which suggests that the ER loss is better to set as equal contributions as the action classification loss.

\end{document}